\PassOptionsToPackage{unicode}{hyperref}
\PassOptionsToPackage{hyphens}{url}
\PassOptionsToPackage{dvipsnames,svgnames,x11names}{xcolor}
\documentclass[
  default,
  sn-basic,
  referee,
  Numbered]{sn-jnl}

\usepackage{xcolor}
\usepackage{amsmath,amssymb}
\usepackage{iftex}
\ifPDFTeX
  \usepackage[T1]{fontenc}
  \usepackage[utf8]{inputenc}
  \usepackage{textcomp} 
\else 
  \usepackage{unicode-math} 
  \defaultfontfeatures{Scale=MatchLowercase}
  \defaultfontfeatures[\rmfamily]{Ligatures=TeX,Scale=1}
\fi
\usepackage{lmodern}
\ifPDFTeX\else
\fi
\IfFileExists{upquote.sty}{\usepackage{upquote}}{}
\IfFileExists{microtype.sty}{
  \usepackage[]{microtype}
  \UseMicrotypeSet[protrusion]{basicmath} 
}{}
\makeatletter
\@ifundefined{KOMAClassName}{
  \IfFileExists{parskip.sty}{%
    \usepackage{parskip}
  }{
    \setlength{\parindent}{0pt}
    \setlength{\parskip}{6pt plus 2pt minus 1pt}}
}{
  \KOMAoptions{parskip=half}}
\makeatother
\makeatletter
\ifx\paragraph\undefined\else
  \let\oldparagraph\paragraph
  \renewcommand{\paragraph}{
    \@ifstar
      \xxxParagraphStar
      \xxxParagraphNoStar
  }
  \newcommand{\xxxParagraphStar}[1]{\oldparagraph*{#1}\mbox{}}
  \newcommand{\xxxParagraphNoStar}[1]{\oldparagraph{#1}\mbox{}}
\fi
\ifx\subparagraph\undefined\else
  \let\oldsubparagraph\subparagraph
  \renewcommand{\subparagraph}{
    \@ifstar
      \xxxSubParagraphStar
      \xxxSubParagraphNoStar
  }
  \newcommand{\xxxSubParagraphStar}[1]{\oldsubparagraph*{#1}\mbox{}}
  \newcommand{\xxxSubParagraphNoStar}[1]{\oldsubparagraph{#1}\mbox{}}
\fi
\makeatother

\usepackage{longtable,booktabs,array}
\usepackage{calc} 
\usepackage{etoolbox}
\makeatletter
\patchcmd\longtable{\par}{\if@noskipsec\mbox{}\fi\par}{}{}
\makeatother
\IfFileExists{footnotehyper.sty}{\usepackage{footnotehyper}}{\usepackage{footnote}}
\makesavenoteenv{longtable}
\usepackage{graphicx}
\makeatletter
\newsavebox\pandoc@box
\newcommand*\pandocbounded[1]{
  \sbox\pandoc@box{#1}%
  \Gscale@div\@tempa{\textheight}{\dimexpr\ht\pandoc@box+\dp\pandoc@box\relax}%
  \Gscale@div\@tempb{\linewidth}{\wd\pandoc@box}%
  \ifdim\@tempb\p@<\@tempa\p@\let\@tempa\@tempb\fi
  \ifdim\@tempa\p@<\p@\scalebox{\@tempa}{\usebox\pandoc@box}%
  \else\usebox{\pandoc@box}%
  \fi%
}
\def\fps@figure{htbp}
\makeatother

\usepackage{graphicx}%
\usepackage{multirow}%
\usepackage{amsmath,amssymb,amsfonts}%
\usepackage{amsthm}%
\usepackage{mathrsfs}%
\usepackage[title]{appendix}%
\usepackage{xcolor}%
\usepackage{textcomp}%
\usepackage{manyfoot}%
\usepackage{booktabs}%
\usepackage{algorithm}%
\usepackage{algorithmicx}%
\usepackage{algpseudocode}%
\usepackage{listings}%

\usepackage{bookmark}
\usepackage{hyperref}
\hypersetup{
  colorlinks=true,
  linkcolor=blue,
  citecolor=blue,
  urlcolor=blue,
  breaklinks=true
}
\DeclareOldFontCommand{\rm}{\normalfont\rmfamily}{\mathrm}
\makeatletter
\@ifpackageloaded{caption}{}{\usepackage{caption}}
\AtBeginDocument{%
\ifdefined\contentsname
  \renewcommand*\contentsname{Table of contents}
\else
  \newcommand\contentsname{Table of contents}
\fi
\ifdefined\listfigurename
  \renewcommand*\listfigurename{List of Figures}
\else
  \newcommand\listfigurename{List of Figures}
\fi
\ifdefined\listtablename
  \renewcommand*\listtablename{List of Tables}
\else
  \newcommand\listtablename{List of Tables}
\fi
\ifdefined\figurename
  \renewcommand*\figurename{Figure}
\else
  \newcommand\figurename{Figure}
\fi
\ifdefined\tablename
  \renewcommand*\tablename{Table}
\else
  \newcommand\tablename{Table}
\fi
}
\@ifpackageloaded{float}{}{\usepackage{float}}
\floatstyle{ruled}
\@ifundefined{c@chapter}{\newfloat{codelisting}{h}{lop}}{\newfloat{codelisting}{h}{lop}[chapter]}
\floatname{codelisting}{Listing}

\makeatother
\makeatletter
\@ifpackageloaded{caption}{}{\usepackage{caption}}
\@ifpackageloaded{subcaption}{}{\usepackage{subcaption}}
\makeatother
\usepackage{bookmark}
\IfFileExists{xurl.sty}{\usepackage{xurl}}{} 
\hypersetup{
  pdftitle={CleanSurvival: Automated data preprocessing for time-to-event models using reinforcement learning},
  pdfauthor={Yousef Koka; David Selby; Gerrit Großmann; Kathan Pandya; Sebastian Vollmer},
  colorlinks=true,
  linkcolor={blue},
  filecolor={Maroon},
  citecolor={Blue},
  urlcolor={Blue},
  pdfcreator={LaTeX via pandoc}}

\title[CleanSurvival: Automated data preprocessing for time-to-event
models using reinforcement learning]{CleanSurvival: Automated data
preprocessing for time-to-event models using reinforcement learning}

\author[1]{\fnm{Yousef} \sur{Koka}}\author*[2,3]{\fnm{David} \sur{Selby}}\email{david.selby@dfki.de}\author[2]{\fnm{Gerrit} \sur{Großmann}}\author*[2,4]{\fnm{Kathan} \sur{Pandya}}\email{kathan.pandya@dfki.de}\author[2,3]{\fnm{Sebastian} \sur{Vollmer}}
\affil[1]{\orgname{German University in Cairo}, \orgaddress{\street{New
Cairo City}, \city{Cairo}, \country{Egypt}}}
\affil[2]{\orgdiv{Data Science and its Applications}, \orgname{German
Research Center for Artificial Intelligence
(DFKI)}, \orgaddress{\street{Trippstadter Str.
122}, \city{Kaiserslautern}, \postcode{67663}, \country{Germany}}}
\affil[3]{\orgname{University of Kaiserslautern--Landau
(RPTU)}, \orgaddress{\street{Gottlieb-Daimler-Str.}, \city{Kaiserslautern}, \postcode{67663}, \country{Germany}}}
\affil[4]{\orgname{University of
Saarland}, \orgaddress{\street{Campus}, \city{Saarbrücken}, \postcode{66123}, \country{Germany}}}

\abstract{\textbf{Background:} Data preprocessing is often paid little
attention in machine learning, despite its potentially significant
impact on model performance. While automated machine learning pipelines
are starting to recognise and integrate data preprocessing into their
solutions for classification and regression tasks, this integration is
lacking for more specialised tasks like time-to-event models for
censored data. As a result, survival analysis not only faces the general
challenges of data preprocessing but also suffers from the lack of
tailored, automated solutions in this area.

\textbf{Method:} To address this gap, this paper presents
\texttt{CleanSurvival}, a reinforcement-learning-based solution for
optimizing preprocessing pipelines, extended specifically for survival
analysis. The framework can handle continuous and categorical variables.
It builds upon Learn2Clean's \(Q\)-learning to select which combination
of data imputation, outlier detection and feature extraction techniques
achieves optimal performance for a Cox, random forest, neural network or
user-supplied time-to-event model. The Python package is available on
GitHub: \url{https://github.com/datasciapps/CleanSurvival}.

\textbf{Results:} Experimental benchmarks on real-world datasets show
that the \(Q\)-learning-based data preprocessing can improve predictive
performance relative to simple baselines, while runtime behaviour is
condition-dependent and most clearly interpretable in the best-covered
benchmark cells. Furthermore, a simulation study demonstrates
effectiveness across different types and levels of missingness and
noise.

\textbf{Conclusion:} With an increase in the use of machine learning, it
becomes important to generalise AutoML pipelines to a variety of models
now present, including survival analysis. Tools like
\texttt{CleanSurvival}, which integrate preprocessing for survival
analysis, can make survival studies faster and easier to perform, while
also yielding more robust results.}

\begin{document}
\maketitle

\section{Introduction}\label{sec-intro}

In the era of big data and machine learning (ML), the ability to extract
meaningful insights from complex datasets is paramount. A critical step
in this process is \emph{data preprocessing}, which involves cleaning,
transforming, and preparing raw data to be suitable for analysis. The
quality of data preprocessing may significantly impact the performance
and reliability of ML models \citep{Cote2024}, even for robust learning
methods \citep{ilyas2022}. This is particularly important in the field
of survival analysis, where the goal is to predict the time until an
event of interest occurs, such as patient death, equipment failure or
customer churn.

Survival analysis poses unique challenges due to the presence of
censored data, where the event of interest has not yet been observed.
However, it is also overlooked in the context of automated machine
learning (AutoML) pipelines, which aim to streamline the ML development
process by automating tasks such as algorithm selection, hyperparameter
tuning and model evaluation---and increasingly incorporate data
preparation as part of their pipelines
\citep{zoeller2021_autoML_benchmark, mumuni_automated_2025, mladenovic2026automated}.
There exist several tools focusing on preprocessing for classification
and regression models
\citep[e.g.][]{learn2clean_2019, bilal2022, nikitin_fedot_2022, santos2023}.
In this paper, we introduce \texttt{CleanSurvival}, an automated data
preprocessing framework tailored specifically for survival analysis.

\texttt{CleanSurvival} leverages reinforcement learning (RL) techniques,
specifically \(Q\)-learning, to optimise decisions such as imputation of
missing values, detection and handling of outliers and feature
extraction for survival models. RL is a powerful approach for automated
data preprocessing because it dynamically optimises pipeline steps by
learning to maximise a reward function tied directly to model
performance, ensuring data cleaning decisions are guided by their impact
on survival predictions. The framework is designed to handle continuous
and categorical variables, and can be used with a variety of
time-to-event models, from classical methods to modern deep learning
frameworks.

The framework, available as an open-source Python package, is
demonstrated on several common survival analysis datasets, highlighting
the sensitivity of predictive performance to data preprocessing steps
and boasting improved predictive performance compared to standard
approaches.

The article is organised as follows. Section~\ref{sec-background}
provides an overview of data preprocessing, survival analysis, AutoML
tools and \(Q\)-learning. Section~\ref{sec-methods} describes the
architecture of \texttt{CleanSurvival} and its features.
Section~\ref{sec-experiments} presents the results of experimental
evaluation of the framework on real-world datasets and
Section~\ref{sec-discussion} discusses CleanSurvival in comparison to
other upcoming tools in this field while Section~\ref{sec-limitations} 
highlights some limitations. Finally, Section~\ref{sec-conclusion} discusses 
the results and outlines future directions for research.

\section{Background}\label{sec-background}

Data preprocessing involves cleaning, transforming and organizing raw
data into a suitable format for analysis, and is an important step in the
ML pipeline. Sub-tasks of data preprocessing include imputation or
removal of missing values, detection and handling of outliers, variable
selection, feature extraction and data transformations. These steps can
have a profound downstream impact on classification performance
\citep{li_cleanml_2021} and model explanations
\citep{sharma_x-hacking_2024}.

However, the selection of appropriate preprocessing methods often
requires a combination of domain knowledge, visual inspection and manual
experimentation; it is also often poorly documented, whether in academic
papers or computational notebooks
\citep{strasser2024, golendukhina2024}. Some authors have even attempted
to quantify the effect of preprocessing steps on model predictions
independently of the dataset \citep{gonzalezzelaya2019}.

Automated data preprocessing has emerged to address these challenges
\citep{bilal2022, santos2023, salhi2023, mumuni_automated_2025}. This
approach uses algorithms and heuristics to automate various data
cleaning and transformation tasks, reducing the need for manual
intervention or iteration and potentially improving the efficiency and
effectiveness of the preprocessing stage, ideally by learning from past
cleaning tasks \citep{mahdavi_towards_2019}. However, the field is still
in its infancy.

\subsection{Survival analysis}\label{survival-analysis}

Survival analysis, also known as reliability analysis or duration
modelling, is a statistical method for analysing time-to event data. It
is widely used in various fields, including medicine, engineering and
social sciences. In survival analysis, the primary goal is to model the
time until an event of interest occurs, such as death, disease
progression, machine failure or customer churn. The survival function,
\[S(t) = P(T > t),\] denotes the probability that the time of event
(death), a random variable \(T\), occurred later than a time \(t\). A
unique characteristic of time-to-event problems is censoring, or data
points that are only partially observed, such as patients who survived
up until the last observation time, at which point they were lost to
follow-up or the study period ended \citep{wang2019}.

Naïvely, one can treat survival analysis as either a regression or
classification problem, but both approaches lead to a significant
information loss. In the former case, one treats the observed survival
time as a continuous outcome, and censored observations are either
discarded or imputed, resulting in substantial reduction in sample size
or bias introduced by oversimplified assumptions about the censoring
process. In the latter case, one models survival---or not---in a
predefined time window, reducing the problem to binary classification
and losing granular information about event times
\citep{leung_censoring_1997}.

Survival analysis models, such as the Cox proportional hazards model,
Kaplan--Meier estimator and accelerated failure time models, are widely
used in practice \citep{wang2019, singh_applications_2022}. These models
estimate the hazard function, survival function, or survival
probabilities over time, providing valuable insights into the
relationship between covariates and survival outcomes.

Concordance indices like the C-index are widely used in the evaluation
of survival models due to their simplicity and ease of interpretation
\citep{longato2020}. The C-index evaluates the model's ability to
correctly rank individuals based on their risk of experiencing the
event. A higher C-index indicates better discriminatory power.

However, concordance indices have significant limitations, due to their
poor calibration and failure to consider the distribution of survival
times as well as their ranks \citep{hartman_pitfalls_2023}. The Brier
score is another metric in survival analysis which calculates the
accuracies of the predicted survival probabilities at a given time point
\citep{Steyerberg2019}. Other calibration metrics like Houwelingen's
\(\alpha\) or D-calibration can also be used to measure how well
survival probabilities align with observed event times
\citep{vanhouwelingen2000}. They could be compared against a baseline
model such as the Kaplan--Meier estimator. Weighted integrated survival
log loss or integrated Brier score (also known as the integrated Graf
score) are recommended scoring rules \citep{sonabend_phd_2021}.

To account for censoring, we apply inverse probability of censoring
weighting (IPCW). Let \(G(t)\) be the Kaplan--Meier estimate of the
probability of not being censored at \(t\). Then, the integrated Brier
score (IBS) is defined \[
\text{IBS} = \frac{1}{n} \sum_{i=1}^n \sum_{j=1}^k \Delta t_j \, \frac{\left( S(t_j | x_i) - \text{Observed}(t_j, i) \right)^2}{ G(t_j) },
\] for a set of time points \(\{t_1, t_2, \ldots, t_k\}\), where
\(\Delta t_j\) represents the difference between time points, \(n\) is
the number of individuals and \(\text{Observed}(t, i)\) is an indicator
function, equal to \(1\) if the individual is known to have survived
beyond \(t\) and \(0\) otherwise.

\subsection{Automated machine
learning}\label{automated-machine-learning}

Automated machine learning (AutoML) aims to streamline the process of ML
development by automating steps such as algorithm selection,
hyperparameter tuning and model evaluation, reducing the amount of time
and expertise required by practitioners to train, deploy and fine-tune
models \citep{feurer_autosklearn_2022, barbudo2023}. AutoML frameworks
have seen success in various applications by employing search strategies
such as meta-learning, Bayesian optimisation and ensemble learning to
achieve competitive performance.

However, data preprocessing remains an important analysis step that
typically falls outside the AutoML pipeline
\citep{paranjape_automated_2022}. Data preparation steps including
cleaning, normalisation and feature engineering are critical for the
success of ML models but can be highly problem-specific
\citep{kuhn_feature_engineering_2019}. Automating these tasks while
maintaining flexibility for diverse datasets remains a significant
hurdle \citep{salhi2023, mumuni_automated_2025}. Mahdavi et al.
\citep{mahdavi_towards_2019} highlighted the potential of AI to solve
data quality problems through data profiling and learning from past
cleaning attempts \citep{mahdavi_semi_2021}. A holistic approach
integrates the cleaning process with downstream tasks so that the
cleaning is optimised for predictive performance
\citep{neutatz_cleaning_2021}; indeed when integrated into an AutoML
framework, some cleaning steps may be more important than others
\citep{neutatz_data_2022}.

Survival analysis in particular faces particular challenges, partly due
to the relative lack of support for such models in the first place
(versus classification or regression), as well as unique difficulties of
handling censored time-to-event data \citep{wang2019}, which are
typically not addressed in conventional AutoML frameworks and do not
feature in AutoML surveys \citep{barbudo2023}. Seamlessly integrating
these domain-specific preprocessing steps with the downstream tasks of
model optimisation and evaluation is a complex, underexplored area.

Prominent AutoML pipelines include \texttt{Auto-WEKA}, an early AutoML
system that uses Bayesian optimisation to search for the best
combination of preprocessing steps and machine learning algorithms
\citep{thornton_autoweka_2013}. It covers a range of models but does not
cover crucial survival models like Cox or survival metrics
\citep{hutter_auto-weka_2019}. \texttt{TPOT} is a tree-based pipeline
optimisation tool that uses genetic programming to evolve pipelines of
data cleaning and machine learning operations, however it performs
simple imputation and does not perform censoring natively, thus
cannot work on survival analysis \citep{olson_tpot_2016}. Similarly,
\texttt{auto-sklearn} (an extension of \texttt{Auto-WEKA} that
incorporates more recent advancements in machine learning and
hyperparameter optimisation, while offering a familiar interface based on
the Python package \texttt{scikit-learn}) cannot perform preprocessing
for survival models and metrics \citep{feurer_autosklearn_2022}.

\citet{salhi2023} presented a recent survey of data preprocessing using
AutoML (though survival analysis is not mentioned). In their review,
they highlight the relative capabilities of AutoML platforms: in many
cases the data processing support is relatively basic. All 11 tools
reviewed support missing value imputation, though for specific
frameworks like \texttt{auto-sklearn}, certain types of missing values
must still be handled manually by the user before the automated pipeline
can proceed. In such cases, claims of support for automated data
processing may reflect the capabilities of the underlying ML framework
(i.e.~\texttt{scikit-learn}) rather than fully autonomous handling.
\citet{mumuni_automated_2025} also surveyed automated data processing
for deep learning applications, highlighting in more detail the extent
to which data processing steps are integral components of the automated
pipeline. They similarly note the lack of early support from
\texttt{auto-sklearn}.

\begin{longtable}[]{@{}
  >{\raggedright\arraybackslash}p{(\linewidth - 10\tabcolsep) * \real{0.1667}}
  >{\raggedright\arraybackslash}p{(\linewidth - 10\tabcolsep) * \real{0.1667}}
  >{\raggedright\arraybackslash}p{(\linewidth - 10\tabcolsep) * \real{0.1667}}
  >{\raggedright\arraybackslash}p{(\linewidth - 10\tabcolsep) * \real{0.1667}}
  >{\raggedright\arraybackslash}p{(\linewidth - 10\tabcolsep) * \real{0.1667}}
  >{\raggedright\arraybackslash}p{(\linewidth - 10\tabcolsep) * \real{0.1667}}@{}}
\caption{Comparison of AutoML frameworks with respect to data
preprocessing and survival analysis
capabilities}\label{tbl-automl-features}\tabularnewline
\toprule\noalign{}
\begin{minipage}[b]{\linewidth}\raggedright
Framework
\end{minipage} & \begin{minipage}[b]{\linewidth}\raggedright
License
\end{minipage} & \begin{minipage}[b]{\linewidth}\raggedright
Missing values
\end{minipage} & \begin{minipage}[b]{\linewidth}\raggedright
Categorical
\end{minipage} & \begin{minipage}[b]{\linewidth}\raggedright
Imbalance
\end{minipage} & \begin{minipage}[b]{\linewidth}\raggedright
Survival
\end{minipage} \\
\midrule\noalign{}
\endfirsthead
\toprule\noalign{}
\begin{minipage}[b]{\linewidth}\raggedright
Framework
\end{minipage} & \begin{minipage}[b]{\linewidth}\raggedright
License
\end{minipage} & \begin{minipage}[b]{\linewidth}\raggedright
Missing values
\end{minipage} & \begin{minipage}[b]{\linewidth}\raggedright
Categorical
\end{minipage} & \begin{minipage}[b]{\linewidth}\raggedright
Imbalance
\end{minipage} & \begin{minipage}[b]{\linewidth}\raggedright
Survival
\end{minipage} \\
\midrule\noalign{}
\endhead
\bottomrule\noalign{}
\endlastfoot
auto-sklearn & Open-source & Requires NaN removal & Yes & Yes & External
libraries \\
TPOT & Open-source & Yes & No & No & No \\
Google AutoML Tables & Proprietary & Yes & Yes & Yes & No \\
H2O AutoML & Open-source & Yes & Yes & Yes & Limited \\
PyCaret & Open-source & Yes & Yes & Yes & Limited \\
AutoKeras & Open-source & Yes & Yes & Limited & Custom models \\
Azure AutoML & Proprietary & Yes & Yes & Yes & External tools \\
BigML & Commercial & Yes & Yes & Limited & No \\
MLflow & Open-source & No & No & No & No \\
FLAML & Open-source & No & No & Yes & No \\
MLJAR & Open-source & Yes & Yes & Limited & No \\
DataRobot & Proprietary & Yes & Yes & Yes & Limited \\
Amazon SageMaker Autopilot & Proprietary & Yes & Yes & Yes & No \\
AutoGluon & Open-source & Yes & Yes & Limited & External libraries \\
\end{longtable}

Table~\ref{tbl-automl-features} compares the features of various AutoML
solutions. Of frameworks offering automated data cleaning, DataRobot is
a proprietary, commercial platform that models time-to-event data by
converting the task into a classification problem via time
discretisation \citep[also called `survival stacking',][]{Craig2025}.
Google AutoML Tables, Amazon SageMaker Autopilot and Azure ML can
perform different preprocessing steps, but they are not always tailored
to survival analysis and these are all proprietary tools
\citep{nayak_sagemaker_2024}. Big ML, which performs decision tree-based
optimisation, enables preprocessing but not survival analysis
\citep{big_ml_features}. MLJAR and AutoGluon both use stack ensembling,
and perform most of the preprocessing steps needed but do not support
survival analysis \citep{mljar, autogluon_survival}. PyCaret can perform
preprocessing for regression or classification, but cannot handle
censored time-to-event data without external tools \citep{pysurvival}.
H2O.ai can run Cox proportional hazards models \citep{h2o_cox} as a
fixed model, but not via its AutoML interface as it requires advanced
platforms for the same.

However, some dedicated data cleaning solutions have been proposed.
\citet{bilal2022} proposed Auto-Prep, an interactive Python-based tool
that recommends data cleaning methods to the user based on application
of candidate techniques and subsequent evaluation using simple
classifiers or regression models. In their accompanying review of data
preprocessing in AutoML, \citet{bilal2022} highlight the widespread lack
of capabilities among existing tools to perform data preprocessing and
feature engineering without manual human intervention. Another Python
package, Atlantic \citep{santos2023}, automates preprocessing steps
including feature engineering and missing value imputation for
supervised learning tasks. The framework identifies the best combination
of steps based on evaluation using tree-based model ensembles.

MLsurvival \citep{zhou2020} described an automated tool for cancer
survival prediction that removes or imputes missing values, selects and
standardises features, trains survival models and then makes
predictions. Unfortunately, neither a full text article nor open source
implementation of the method were published. More recently,
\citet{pomsuwan2024} proposed an AutoML system for survival analysis
based on genetic algorithms and a combination of elastic-net Cox models,
random survival forests and survival trees, optimizing for C-index.
However, the tool does not incorporate data preprocessing. Other
frameworks like AutoPrognosis \citep{alaa_autoprognosis_2018} have
developed automated tools using Bayesian optimisation that include
survival analysis, though with less scope for customisation of
preprocessing pipelines. The widely used library
\texttt{scikit-survival} integrates \texttt{scikit-learn} preprocessing
with survival models, but it requires manual pipeline building
\citep{community_scikit-survival_2025}. The R package \texttt{mlr3proba}
provides a machine learning interface for survival analysis and
preprocessing, however missingness is dealt with externally
\citep{sonabend_mlr3proba_2021}.

\subsection{\texorpdfstring{\(Q\)-learning}{Q-learning}}\label{q-learning}

Reinforcement learning is well-suited to the task of automating
constrained ML pipelines, as it optimises sequential decision-making
processes, balancing exploration and exploitation, while being less
computationally intensive than other methods, such as unconstrained
evolutionary algorithms \citep{heffetz2020}.

\(Q\)-learning \citep{watkins_qlearning_1992} is a model-free off-policy
reinforcement learning algorithm that seeks to find the optimal
action-selection policy for an agent interacting with an environment.
The algorithm is based on estimating the value \(Q=Q(s, a)\) of taking
an action \(a\) in a given state \(s\). The agent iteratively updates
these \(Q\) values based on its experiences, enabling it to learn an
optimal policy even in environments with stochastic rewards and
transitions.

The goal of \(Q\)-learning is to maximise the cumulative reward over
time by updating \(Q\) according to the Bellman equation. Given a
current state \(s\), action \(a\), reward \(r\) and next state \(s'\),
the update rule is:
\begin{equation}\protect\phantomsection\label{eq-q_learning}{
Q(s, a) \leftarrow Q(s, a) + \alpha \bigl( r + \gamma \max_{a'} Q(s', a') - Q(s, a) \bigr),
}\end{equation} where \(\alpha\) is the learning rate and \(\gamma\) is
the discount factor, controlling the importance of future rewards. The
update equation Equation~\ref{eq-q_learning} allows the \(Q\)-learning
agent to converge to an optimal policy \(\pi^*\), defined
\begin{equation}\protect\phantomsection\label{eq-optimal-policy}{
\pi^*(s) = \arg\max_a Q(s, a).
}\end{equation} without requiring a model of the environment's dynamics.
By exploring various state--action pairs and refining \(Q\)-values,
\(Q\)-learning is able to asymptotically approach optimal behaviour,
provided that the agent balances exploration and exploitation
effectively.

Alternatives to \(Q\)-learning, such as Bayesian optimisation, can offer
improved sample efficiency in some cases but often struggle to scale in
high-dimensional or discrete search spaces typical of complex AutoML
pipelines \citep{learn2clean_2019, gonzales_bayesian_2024}. More
specialised methods, such as neural architecture search, are not easily
extensible to data preprocessing tasks. Other reinforcement learning
approaches, including deep reinforcement learning \citep{heffetz2020}
and Monte Carlo tree search
\citep{drori2019automaticmachinelearningpipeline}, provide flexibility
but introduce additional computational overhead and may require
constraints or tailored mechanisms to ensure valid ML pipelines.

Berti-Equille \citep{learn2clean_2019} developed \texttt{Learn2Clean}, a
tool offering an innovative approach to data preprocessing. It leverages
\(Q\)-Learning to dynamically select the optimal sequence of
preprocessing tasks for a given dataset and ML model. This optimisation
aims to maximise the quality of the ML model's results.
\texttt{Learn2Clean} implements automated preprocessing for regression,
classification and clustering tasks, using RL to optimise respective
evaluation criteria: mean squared error, accuracy and silhouette index.
However, the framework lacks built-in support for time-to-event data,
categorical data types, flexible hyperparameter tuning and custom reward
functions. It also has a complex dependency structure, which can make
initial setup challenging for end-users. We build upon
\texttt{Learn2Clean}, by adapting the reinforcement learning framework
to survival analysis.

\section{Methodology}\label{sec-methods}

\begin{figure*}

\centering{

\pandocbounded{\includegraphics[keepaspectratio]{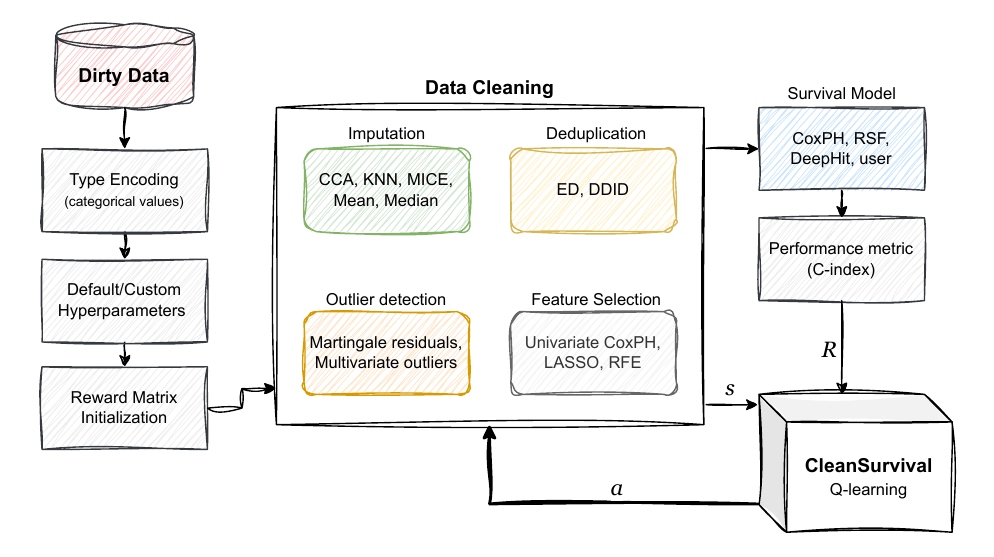}}

}

\caption{\label{fig-architecture}Architecture of the
\texttt{CleanSurvival} automated data cleaning framework}

\end{figure*}

In this section we describe \texttt{CleanSurvival}, our proposed AutoML
data preprocessing tool for survival analysis, illustrated in
Figure~\ref{fig-architecture}.

\subsection{Data preprocessing
methods}\label{data-preprocessing-methods}

\subsubsection{Missing values}\label{missing-values}

\texttt{CleanSurvival} offers a variety of methods for handling missing
values, addressing different data characteristics and analytical goals.

\begin{itemize}
\item
  For straightforward scenarios with minimal missingness, the complete case
  analysis (CCA) simply removes rows containing missing values.
\item
  Simple mean/median imputation replaces missing values with the mean or
  median of the observed values for a given variable.
\item
  Multiple imputation using chained equations \citep{buuren2011}
  provides a more robust approach by generating multiple suitable
  replacements for each missing value, creating several complete
  datasets for analysis. This method utilises an iterative imputer,
  which starts with an initial guess and refines the estimates until
  convergence.
\item
  Finally (and most computationally intensive): \(k\)-nearest neighbors
  (KNN) imputation identifies the \(k\) most similar observations to the
  one with missing values, based on other, non-missing features, and
  uses their values to impute the missing data. Mean or median
  imputation is a limit case where \(k\rightarrow n\).
\end{itemize}

\subsubsection{Outliers}\label{outliers}

To ensure the reliability and validity of survival analysis, robust
outlier detection methods are essential
\citep{carrasquinha_consensus_2018}.

\begin{itemize}
\item
  For multivariate datasets, \texttt{CleanSurvival} employs the Elliptic
  Envelope algorithm \citep{rousseeuw1999} to identify and remove
  outliers based on their Mahalanobis distances from the data centre.
  This method is particularly useful for detecting outliers that deviate
  from the overall correlation structure of the data.
\item
  The Martingale residuals method \citep{therneau1990} calculates the
  difference between the observed and expected number of events for each
  individual (based on a simple Kaplan--Meier estimator), providing a
  measure of how unusual their survival time is compared to the expected
  survival time.
\end{itemize}

\subsubsection{Variable selection and feature
extraction}\label{variable-selection-and-feature-extraction}

To identify the most salient variables for survival analysis, a variety
of feature selection methods are available, enhancing both the model
performance and interpretability.

\begin{itemize}
\item
  The Univariate Cox Proportional Hazards Selection (UC) method assesses
  the individual effect of each feature on survival using the Cox
  Proportional Hazards model. It selects features based on the
  significance of their coefficients, highlighting the variables strongly
  associated with survival outcomes.
\item
  The LASSO (Least Absolute Shrinkage and Selection Operator) regression
  technique shrinks the coefficients of less important features to zero,
  effectively performing feature selection.
\item
  Recursive Feature Elimination (RFE) recursively removes the least
  important features based on their contribution to a model's
  performance, using cross-validation to evaluate the model's
  performance at each step.
\item
  The Information Gain Selection (IG) method measures the amount of
  information gained about the target variable (survival outcome) by
  knowing the value of a feature. This helps identify the most relevant
  variables by selecting features that provide the most information
  about the survival outcome.
\end{itemize}

\subsection{Survival analysis}\label{survival-analysis-1}

Three survival analysis models were carefully selected to integrate into
\texttt{CleanSurvival}, each chosen for its unique strengths and
applicability to a variety of survival analysis scenarios.

\begin{description}
\item[Cox proportional hazards]
This widely-used model \citep{Davidson-Pilon2019} is valued for its
interpretability, allowing researchers to quantify the impact of
different factors on the hazard rate. The core assumption of this model
is that the hazard ratio between any two individuals is constant over
time (the proportional hazards assumption). While violations of this
assumption can lead to biased or uninterpretable estimates in standard
analysis, our pipeline search explicitly optimises for the downstream
C-index. Consequently, the chosen preprocessing transformations (such as
feature selection or imputation) may indirectly help mitigate severe
violations.
\item[Random survival forest]
The RSF model \citep{sksurv} is an ensemble method based on decision
trees, offering robustness to nonlinearities and interactions in the
data. Its non-parametric nature makes it a flexible choice when the
underlying relationships in the data are not well understood.
\item[DeepHit neural network]
This deep learning model \citep{lee2018} leverages the power of neural
networks to capture complex patterns and interactions in survival data.
It is a specialised form of deep neural network, which, instead of a
single risk score, estimates a discrete probability mass function over
fixed time intervals. Its ability to model multiple competing risks
makes it particularly well-suited for scenarios where individuals may
experience different types of events.
\end{description}

\subsubsection{Reward structure}\label{reward-structure}

To guide the \(Q\)-learner effectively for survival analysis problems,
the reward structure uses the concordance index (C-index) on a hold-out
set \citep{longato2020}. Because this reward is evaluated directly on
the downstream survival model's predictions, the \(Q\)-learning agent
identifies preprocessing pipelines that are optimised for the chosen
model class. This means that a preprocessing pipeline that is optimal
for a Cox proportional hazards model may differ from that optimised for
a random survival forest.

\subsection{Working modes}\label{working-modes}

To provide users with a range of options for data preprocessing and
analysis, four distinct working modes were implemented in
\texttt{CleanSurvival}:

\begin{description}
\item[Main algorithm]
This mode uses the core \(Q\)-learning algorithm to identify the optimal
sequence of preprocessing steps that maximises the performance of the
chosen survival analysis model. This is the primary mode for automated
pipeline optimisation.
\item[Random cleaning]
In this mode, users can specify the desired number of random
experiments. The tool generates random preprocessing pipelines and
evaluates their performance, providing insights into the impact of
different preprocessing choices. This mode can serve as a baseline for
comparison with the optimised pipeline.
\item[Custom pipeline]
This mode allows users to define their own fixed preprocessing pipelines
using a simple text configuration file. Each line in the file specifies
a sequence of preprocessing methods, providing flexibility for testing
specific hypotheses or domain knowledge.
\item[No preparation]
This mode bypasses all preprocessing steps, directly passing the raw
dataset to the chosen survival analysis model. This can be useful for
establishing a baseline for performance without any preprocessing.
\end{description}

The inclusion of these working modes significantly enhances the utility
of the framework by offering baselines for evaluating the effectiveness
of the optimised pipelines generated by the \(Q\)-learning algorithm.

\section{Experiments}\label{sec-experiments}

\subsection{Experimental setup}\label{experimental-setup}

All methods were implemented in Python. Source code and documentation
are available at \url{https://github.com/datasciapps/CleanSurvival}.

Details of the datasets used are provided in Table~\ref{tbl-datasets}.
We demonstrate the approach using the \texttt{rotterdam} dataset,
derived from the Rotterdam Study \citep{Royston_Rotterdam_2013}, a large
prospective cohort study in the Netherlands, \texttt{flchain}
\citep{Dispenzieri_FLchain_2012}, a study of the relationship between
serum free light chain (a type of blood measurement) and mortality, and
\texttt{gbsg} \citep{Schumacher1994}, containing data from the German
Breast Cancer Study Group. These datasets are distributed as part of the
R package \texttt{survival} \citep{survival_R, R_core_team2025}. The
\texttt{flchain} dataset contains naturally occurring missing values,
for the `creatinine' and `chapter' variables of 17.15\% and 72.45\%
respectively, while the \texttt{rotterdam} and \texttt{gbsg} datasets do
not contain any missing values, allowing us to simulate different types
and levels of missingness for benchmarking purposes.

\begin{longtable}[]{@{}llll@{}}
\caption{Summary of datasets used in the
experiments}\label{tbl-datasets}\tabularnewline
\toprule\noalign{}
Dataset & Samples & Features & Source \\
\midrule\noalign{}
\endfirsthead
\toprule\noalign{}
Dataset & Samples & Features & Source \\
\midrule\noalign{}
\endhead
\bottomrule\noalign{}
\endlastfoot
Rotterdam & 2982 & 14 & \citet{Royston_Rotterdam_2013} \\
Flchain & 7874 & 11 & \citet{Dispenzieri_FLchain_2012} \\
GBSG & 686 & 8 & \citet{Schumacher1994} \\
\end{longtable}

The results of data preprocessing strategies suggested by
\texttt{CleanSurvival} are compared against the following methods:

\begin{enumerate}
\def\labelenumi{\arabic{enumi}.}
\item
  Complete case analysis (CCA): effectively ignoring the problem of
  missingness and outliers
\item
  Random selection of imputation methods: simulating a grid search over
  possible analysis choices
\item
  Mean imputation, as an example of a reasonable baseline used by an
  analyst not exploring the sensitivity of the model to different
  imputation strategies
\end{enumerate}

We evaluate each method based on predictive performance metrics
(C-index, IBS) as well as the computational time taken to retrieve a
performant pipeline (defined as reaching within 5\% of the final
best-observed C-index for that dataset).

For model evaluation, the current Cox and random survival forest
wrappers use an 80:20 hold-out split (\texttt{test\_size\ =\ 0.2}).

Hyperparameters that are not part of the \texttt{CleanSurvival} search
space were left at their default values for the \texttt{scikit-survival}
models: the Cox model was fit with default penalisation and the RSF was
trained with 100 estimators and a minimum of 3 samples per leaf. The
architectural choices for the DeepHit deep learning framework were as
follows: the architecture consists of a shared sub-network with two
hidden layers, each containing 100 neurons, using ReLU as the activation
function. This is followed by cause-specific sub-networks, each
containing two hidden layers with 50 neurons per layer. Dropout
regularisation was applied to all layers with a keep probability of 0.8
to mitigate overfitting. The model was trained using the Adam optimiser
with a learning rate of 0.001 for 10,000 iterations. In practice, these
options are customizable by the user or could be incorporated into the
reinforcement learning action space as additional modules.

The Rotterdam dataset does not contain missing values, so synthetic
missingness was introduced using the
\href{https://pypi.org/project/jenga/\#description}{Jenga framework}
\citep{Schelter2021}. Three missingness mechanisms were simulated across
multiple variables: missing completely at random (MCAR, where
missingness is independent of any variable), missing at random (MAR,
where missingness depends only on fully observed variables like age or
treatment group), and missing not at random (MNAR, where missingness
depends on the unobserved value itself, simulating the unrecorded worst-case
outcomes). We evaluated these mechanisms at missingness levels ranging
from 10\% to 50\%. After generating pipelines using
\texttt{CleanSurvival} and random search, we computed the integrated Brier
scores (IBS) across dataset variations.

\subsection{Results}\label{results}

To evaluate optimisation dynamics and calibration clearly, we focus on
pipelines optimised for the Cox proportional hazards model. This keeps
convergence and proportional-hazards diagnostics interpretable within a
single model family. DeepHit results are not shown because intensive
neural architecture search/training inside the reinforcement-learning
loop is beyond the scope of this paper.

\begin{table}

\caption{\label{tbl-results}Predictive performance of survival models,
measured by C-index, for 50\% missingness benchmark configurations of
Rotterdam/GBSG and natural missingness for Flchain. Mechanism indicates
the type of synthetic missingness missing at random, missing completely
at random or missing not at random. Entries report mean C-index \(\pm\)
standard error over 50+ runs.}

\centering{

\begin{tabular}{llrrrrr}
\toprule
Dataset & Mechanism & CCA & CleanSurvival & Mean & Optuna & Random\\
\midrule
Flchain & None & 0.395 $\pm$ 0.011 & 0.660 $\pm$ 0.001 & 0.617 $\pm$ 0.002 & 0.656 $\pm$ 0.000 & 0.737 $\pm$ 0.020\\
Rotterdam & MAR & 0.772 $\pm$ 0.005 & 0.811 $\pm$ 0.002 & 0.787 $\pm$ 0.004 & 0.565 $\pm$ 0.051 & 0.828 $\pm$ 0.002\\
Rotterdam & MCAR & 0.789 $\pm$ 0.004 & 0.824 $\pm$ 0.003 & 0.793 $\pm$ 0.004 & 0.614 $\pm$ 0.041 & 0.824 $\pm$ 0.001\\
Rotterdam & MNAR & 0.789 $\pm$ 0.004 & 0.810 $\pm$ 0.002 & 0.783 $\pm$ 0.004 & 0.625 $\pm$ 0.046 & 0.858 $\pm$ 0.007\\
GBSG & MNAR & -- & 0.632 $\pm$ 0.004 & -- & 0.650 $\pm$ 0.000 & 0.684 $\pm$ 0.008\\
\bottomrule
\end{tabular}

}

\end{table}%

Table~\ref{tbl-results} compares survival models fitted to GBSG, Flchain
and the three 50\% Rotterdam missingness settings. In these summaries,
CleanSurvival remains stronger than the simple analyst baselines (CCA
and mean imputation) in the main Rotterdam 50\% settings and on Flchain.

To verify that this pattern is not specific to Cox proportional hazards
models, results for RSF models in similar scenarios are provided
separately in the supplementary material.

Feature-selection operators were frequently selected by the Q-learning
policy: LASSO appeared in 74/109 (67.9\%) CleanSurvival pipelines and
RFE appeared in 46/109 (42.2\%). This suggests that CleanSurvival
achieves performance gains through a policy of dynamically filtering out
noisy features to stabilise the downstream survival models.

\begin{figure}

\centering{

\pandocbounded{\includegraphics[keepaspectratio]{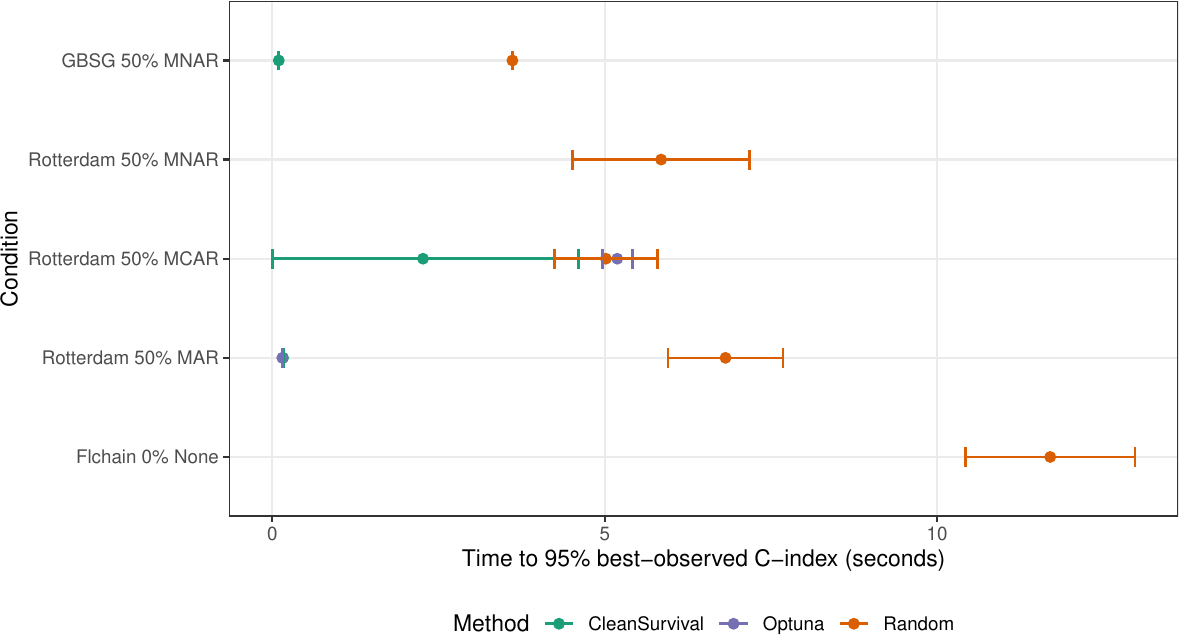}}

}

\caption{\label{fig-runtime}Runtime to target (time to reach 95\% of
best-observed C-index) across selected benchmark settings. Points show
medians, horizontal bars show IQR.}

\end{figure}

In Flchain, random search occasionally found a better edge case, as it
is a high-variance unconstrained exploration, compared to the more
narrowly controlled search of CleanSurvival, which has significantly
reduced runtime as shown in Figure~\ref{fig-runtime}.

\begin{table}

\caption{\label{tbl-paired-inference-compact}Wilcoxon summary
(CleanSurvival vs Optuna/Random) for the selected C-index benchmark.
Rotterdam and GBSG rows correspond to 50\% missingness settings (by
synthetic missingness mechanism: missing at random, missing completely
at random or missing not at random), while Flchain uses its natural
missingness. P-values are Holm-adjusted across all evaluated
conditions.}

\centering{

\begin{tabular}{lllrrr}
\toprule
Dataset & Mechanism & Comparator & $\Delta$ & Win rate & $p$\\
\midrule
Flchain & None & Optuna & -0.00 & 0.48 & 1.49e-02\\
Flchain & None & Random Search & -0.01 & 0.39 & 4.04e-02\\
Rotterdam & MAR & Optuna & 0.80 & 0.68 & 3.54e-05\\
Rotterdam & MAR & Random Search & -0.02 & 0.29 & 3.42e-03\\
Rotterdam & MCAR & Optuna & 0.80 & 1.00 & 2.18e-08\\
\addlinespace
Rotterdam & MCAR & Random Search & -0.00 & 0.50 & 2.29e-01\\
Rotterdam & MNAR & Optuna & 0.80 & 0.66 & 7.30e-05\\
Rotterdam & MNAR & Random Search & -0.03 & 0.22 & 1.20e-05\\
GBSG & MNAR & Optuna & -0.02 & 0.02 & 6.14e-09\\
GBSG & MNAR & Random Search & -0.04 & 0.03 & 3.94e-04\\
\bottomrule
\end{tabular}

}

\end{table}%

In Table~\ref{tbl-paired-inference-compact}, we report Wilcoxon
signed-rank contrasts for paired experimental runs.

Figure~\ref{fig-runtime} summarises runtime behaviour. The complete
numeric runtime table summarizing the distributions across conditions is
provided in the supplementary material.

\begin{figure*}

\centering{

\pandocbounded{\includegraphics[keepaspectratio]{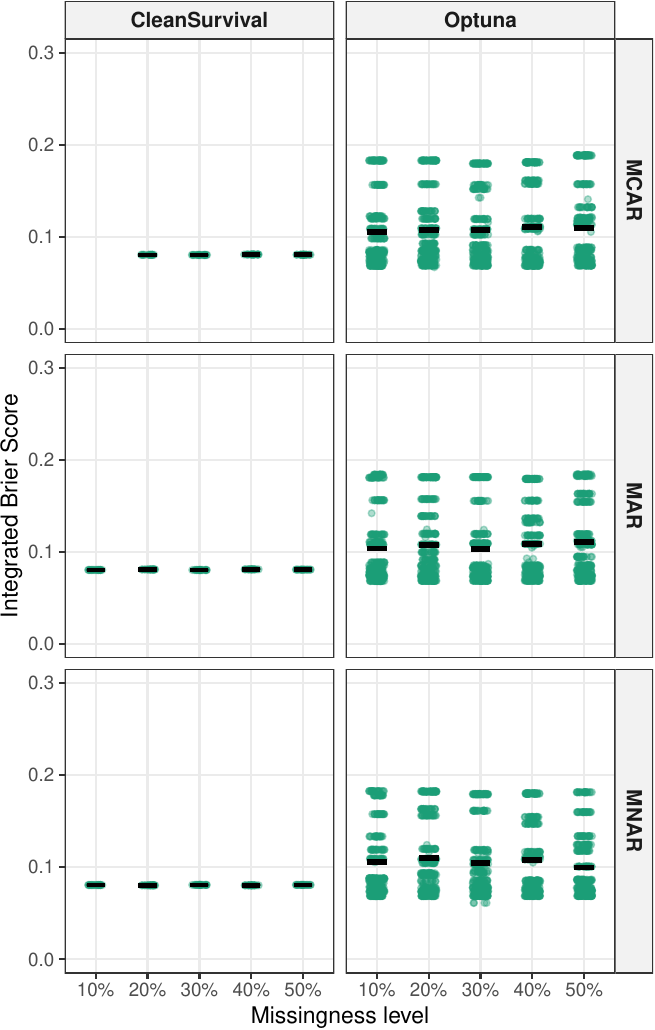}}

}

\caption{\label{fig-ibs-comparison}Integrated Brier Score (IBS) for
Q-learning and Bayesian optimisation on Cox model pipelines with the
Rotterdam dataset across different missingness structures. Each point
represents a single evaluated pipeline. Crossbars show mean ± SE. Lower
is better.}

\end{figure*}%

Figure~\ref{fig-ibs-comparison} demonstrates that CleanSurvival yields
highly stable, low-variance performance in integrated Brier Score across
all missingness structures. In contrast, Optuna exhibits significant
variance apparently due to unstable calibration in unconstrained
pipeline optimisation.

\begin{figure*}

\centering{

\pandocbounded{\includegraphics[keepaspectratio]{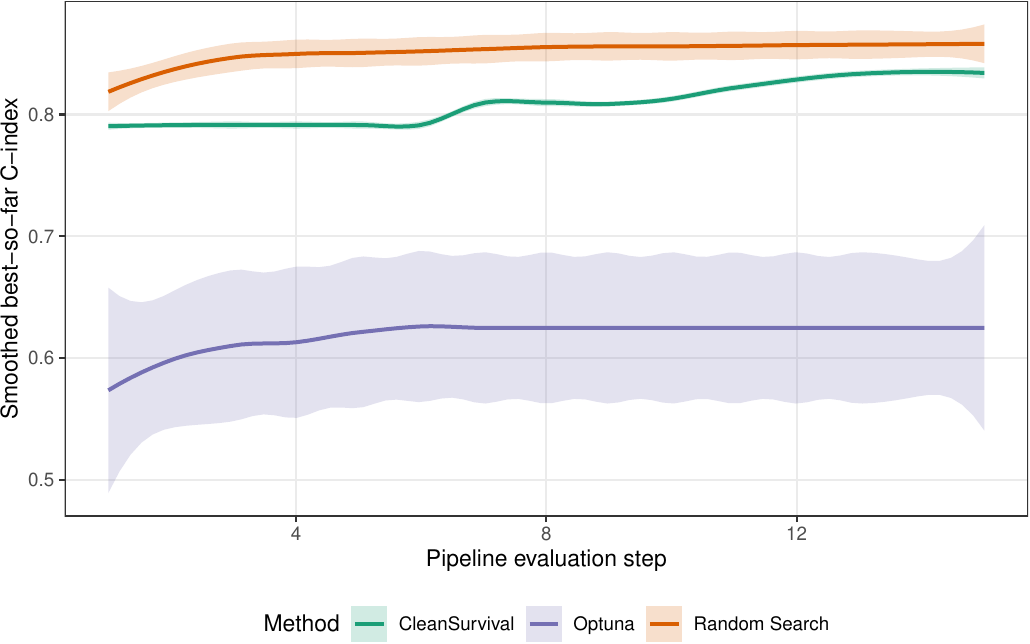}}

}

\caption{\label{fig-grid-search}Aggregate convergence (step) for
Rotterdam at 50\% MNAR. Curves show LOESS-smoothed best-so-far C-index
trajectories for CleanSurvival, Optuna, and Random Search.}

\end{figure*}%

Figure~\ref{fig-grid-search} shows the step-based aggregate convergence
(the average best-historical C-index at each sequential RL iteration).
Traces are smoothed using locally estimated scatterplot smoothing
(LOESS) to illustrate the learning trajectory.

\begin{figure*}

\centering{

\pandocbounded{\includegraphics[keepaspectratio]{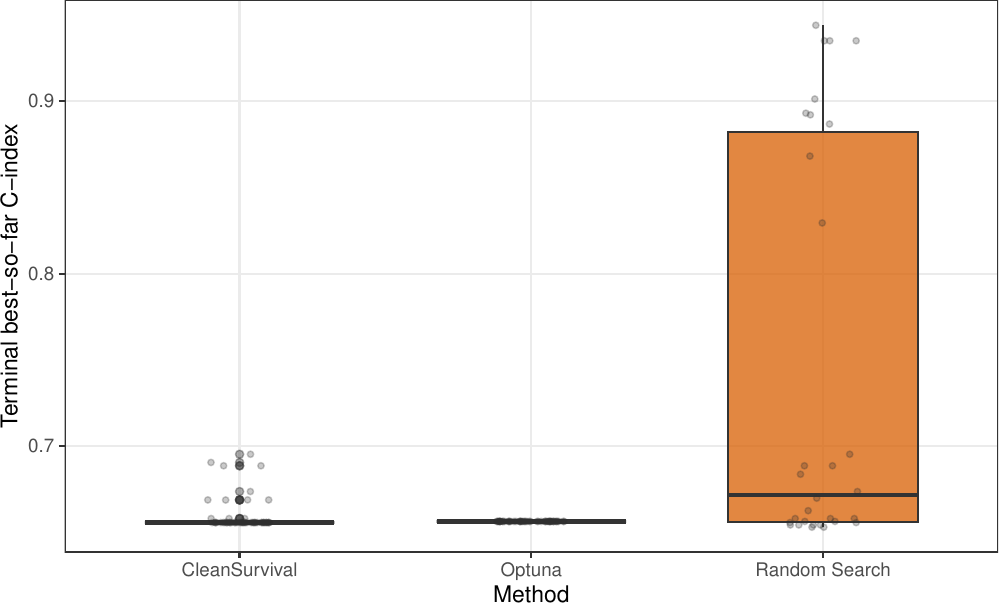}}

}

\caption{\label{fig-flchain-terminal-boxplot}Algorithm terminal
performance on Flchain (C-index). Each box summarises the per-run
terminal best-so-far C-index; unknown method labels are excluded.}

\end{figure*}%

Figure~\ref{fig-flchain-terminal-boxplot} summarises the terminal
best-so-far C-index per run, excluding a marginal fraction of runs
compromised by telemetry logging errors.

\begin{figure*}

\centering{

\pandocbounded{\includegraphics[keepaspectratio]{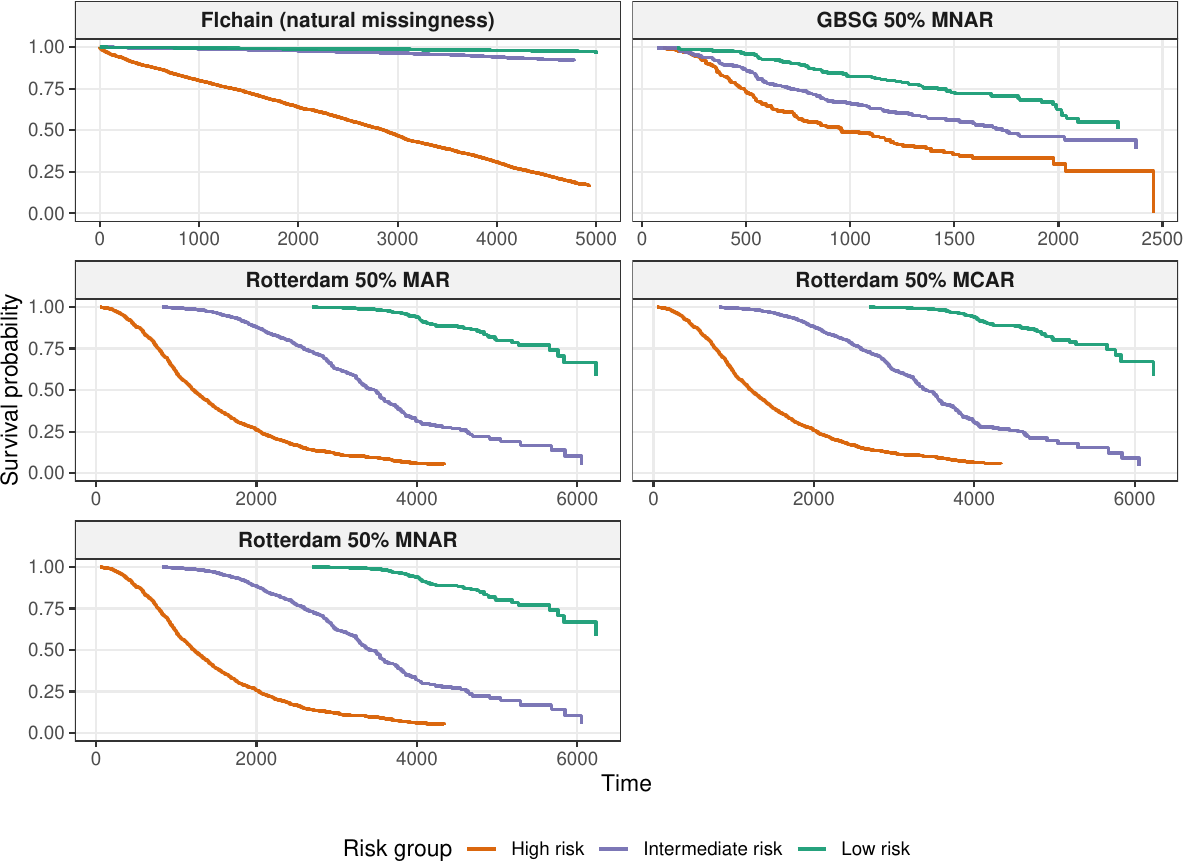}}

}

\caption{\label{fig-km-risk-strata}Kaplan--Meier survival curves by
model-derived risk stratum across key benchmark conditions. For each
condition, a baseline Cox model is fit using simple median imputation
applied within that specific missingness scenario, subjects are divided
into low/intermediate/high risk tertiles by linear predictor, and
observed Kaplan--Meier curves are plotted per stratum.}

\end{figure*}%

Figure~\ref{fig-km-risk-strata} demonstrates CleanSurvival's clinical
utility. By successfully stratifying patients into distinct low,
intermediate, and high-risk groups, it shows that the automated
preprocessing preserves underlying survival signals. Diagnostic plots of
Schoenfeld residuals are provided in the supplementary material.

\section{Discussion}\label{sec-discussion}

With most of the AutoML preprocessing tools focusing on classification
and regression models, we present CleanSurvival as a tool which builds
upon Learn2Clean to tailor preprocessing for survival analysis using a
\(Q\)-learning-based framework and reward structures. Survival analysis
is important in fields like cancer research, and with the abundance of
machine learning algorithms, it becomes crucial to develop automated
approaches that can handle the unique challenges of censored
time-to-event data \citep{Tizi2023}. Our results demonstrate that
integrating data cleaning directly into the optimisation loop---guided
by survival-specific predictive performance metrics---yields pipelines
that are adapted to the downstream model.

The Integrated Brier Score (IBS) evidence indicates that calibration is
broadly stable across most missingness settings for the Cox model, with
the exception of the 40\% MCAR condition, which exhibited higher
variance.

This makes CleanSurvival a valuable flexible tool in the emerging field
of automation for machine learning and preprocessing pipelines, capable
of supporting various survival models, including neural networks, while
permitting user customisation.

In a realistic, practical scenario---such as when analysing
high-dimensional electronic health records---practitioners often
struggle to choose the best sequence of imputation, transformation, and
feature selection strategies \citep{VafaeiSadretal2025}. Rather than
manually testing various missingness imputation rules or arbitrary
variable selections, a clinical data scientist can configure
\texttt{CleanSurvival} with their chosen downstream analysis method
alongside the provided dataset. The framework automates the
hyperparameter combinations and step-order decisions. The optimised
outputs are fully compatible with external libraries; the extracted data
transformations from the final best pipeline can be seamlessly
serialised and utilised within \texttt{scikit-survival}'s scikit-learn
compatible wrapper or passed on to subsequent downstream architectures
like those found in AutoPrognosis.

Furthermore, integrating diagnostic tools early in the automated process
is necessary for clinical reporting. It allows quicker analysis of data
and can allow better translation of results into clinical practice
\citep{boyce2017integrating}. As outlined in the methods and empirical
derivations in the Appendix, the optimisation process itself tends to
select pipelines which naturally stabilise model assumptions on average,
such as keeping proportional hazards relatively constant for the Cox
model. By filtering noisy features out via feature reduction,
CleanSurvival inherently guides the final model towards stability in
complex, high-missingness regimes.

\section{Limitations}\label{sec-limitations}

Several limitations must be acknowledged. First, although the analysis
includes three datasets, the generalisation of these performance gains
to higher-dimensional or multi-modal clinical datasets remains
unverified. Second, the current implementation has computational
overhead, depending on the configurations selected, especially when
utilizing \(k\)-nearest neighbors imputation and complex deep learning
survival models repeatedly alongside the \(Q\)-learning iterations.
Isolating the marginal benefits of specific search-space subsets via
ablation studies is left for future work.

\section{Conclusion}\label{sec-conclusion}

In this paper, we have introduced \texttt{CleanSurvival}, a
\(Q\)-learning-based framework that automates data preprocessing for
survival analysis, addressing the often underserved challenge of
automation in the presence of censored time-to-event data. We
demonstrated its ability to learn performant preprocessing pipelines
across different settings, compared to complete case analysis, mean
imputation, and undirected random search.

The framework's adaptability to various survival analysis models and its
support for diverse preprocessing techniques make it a valuable tool for
researchers and practitioners. Experimental results confirm that
\texttt{CleanSurvival} can facilitate the discovery of optimal pipelines
and maintain robust performance under different missingness patterns,
though comparative runtimes remain heavily condition-dependent. These
findings underscore the critical role of automated data preprocessing in
enhancing the reliability of survival models and the feasibility of
integrating reinforcement learning techniques into AutoML workflows.

Future work will focus on extending the framework to handle additional
preprocessing tasks, incorporating advanced reinforcement learning
strategies and improving scalability for large datasets and complex
pipelines. Additionally, whether ensembling over differently cleaned
datasets, together with raw data, can increase predictive performance or
instead increase data-processing bias, is an open question for future
research.

\section{Declarations}\label{sec-declarations}

\subsection{Ethics approval and consent to
participate}\label{ethics-approval-and-consent-to-participate}

Not applicable. This study used publicly available datasets.

\subsection{Consent for publication}\label{consent-for-publication}

Not applicable.

\subsection{Availability of data and
materials}\label{availability-of-data-and-materials}

All datasets and code are available at
\url{https://github.com/datasciapps/CleanSurvival}.

\subsection{Competing interests}\label{competing-interests}

The authors declare no competing interests.

\subsection{Funding}\label{funding}

This work was supported by the German Federal Ministry of Research,
Technology and Space Travel (BMFTR) via the project ``Eventful'' under
grant ID \texttt{01IW23005}.

\subsection{Authors' contributions}\label{authors-contributions}

YK conceptualised the package, wrote and implemented the code and wrote
the manuscript. DS provided supervision, assisted in conceptualizing,
writing and editing the manuscript. GG helped in code optimisation and
implementation. SV provided supervision, funding acquisition and helped
in reviewing and editing the manuscript. KP performed data evaluation,
code implementation and editing of the manuscript.

\subsection{Acknowledgement}\label{acknowledgement}

We are grateful for the input of Sergey Redyuk, and the helpful comments
of the anonymous reviewers, all of which have greatly improved the
manuscript.

\bibliography{references.bib}

\end{document}